# Spiking Neural Network Predicting Sequence of the External Worlds States in Model-Based Reinforcement Learning

A Preprint

Mikhail Kiselev[0000-0001-7403-6418]

Chuvash State University, Cheboxary, Russia
mkiselev@chuvsu.ru

September 1, 2026

**Abstract.** This paper presents a spiking neural network (SNN) designed to predict the sequence of the external world states starting from the current world state. This SNN does not create the world dynamics model – instead it incorporates the SNN trained to predict the next world state and provides all mechanisms necessary to make the chain of predicted world states. These mechanisms are entirely spiking – they are implemented as spiking neuron ensembles. The present article describes this neuronal structure and tests its operation on a classic RL benchmark – ATARI ping-pong.

## 1 Introduction

This article represents the next step to entirely spiking implementation of model-based reinforcement learning (RL) which still remains a challenging and topical scientific problem. In our previous work [1] we invented SNNs capable of learning an internal representation of the external world dynamics model. The present paper is devoted to application of this model to prediction of chain of subsequent world states – starting from the current one. It is important because such an application of this model is the primary goal why it has been created.

The short review of recent works devoted to spiking model-based RL can be found in [1]. For convenience, I copy it here.

While efficient implementations of model-based RL in traditional artificial neural networks (ANNs) have been reported in several works [2–4], the author is not aware of any its successful implementation in SNNs — if to say only about the projects where all substantial information processing logics are implemented inside SNN, without involvement of outside digital processing modules. This can be attributed to several factors, but primarily to the discrete nature of SNNs, which presents a major obstacle to directly applying the error backpropagation principle in the spiking domain. Backpropagation is a highly efficient and well-studied algorithm applicable to many neural network learning tasks, including RL. However, it requires the loss function to be a smooth, differentiable function of the synaptic weights, which is not the case for SNNs. Therefore, it is required either to modify the backpropagation algorithm making it applicable to SNN or to develop completely different learning principles. The first way seems easier and was selected by many research groups. Most commonly, the surrogate gradient technique [5, 6] is used, where the discrete behavior of spiking neurons is approximated by smooth functions. This allows for the calculation of pseudo-gradients, enabling error backpropagation. However, this approach has many drawbacks. SNNs created using this principle are usually order of magnitude larger than their ANN prototypes. In real-world RL tasks, time is crucial—exact timing of external events matters, and agent actions must span specific time intervals. However, the backpropagation algorithm itself does not incorporate the concept of time. Besides that, while SNNs are most efficient when implemented in specialized neuroprocessors like Intel's Loihi, there are significant problems with implementation of backpropagation on this kind of hardware. Finally, SNNs are regarded as more biologically plausible models than traditional neural networks, which descend from perceptrons. This biological plausibility is expected to give SNNs advantages similar to those of the living brain, such as low energy consumption and few-shot or incremental learning. However, no equivalent of the backpropagation algorithm has been found in the brain.

For this reason, much attention in the world of SNN is paid to the so-called local learning algorithms as an alternative to backpropagation. Until now, the works devoted to SNN-based RL, in which the local synaptic plasticity rules are used, are relatively few. And almost all of them are focused on model-free RL [7 - 10]. SNN implementation of model-based RL remains a largely unexplored area. The work [11] describes an SNN that solves model-based RL problems. However, a significant part of the system (the arbitration component) is implemented outside the network – so that it cannot be called a purely SNN solution. The mechanism used by SNN to learn rules of the external world dynamics in terms of hidden Markov states is described in [12], but it is based on a very simple discrete formalization of external world dynamics which does not include temporal aspect and, therefore, can hardly be used in real-world problems. A similar problem is addressed in the recent research [13]. However, the world dynamics are treated there as a sequence of discrete states, with the SNN's goal being to predict the next state in the sequence—bypassing questions such as when the system will transition to the next state or how long it will remain in the current one. Thus, to the best of our knowledge, our work described in [1] is the first to demonstrate the ability of SNNs to learn external world dynamics from scratch, using only spiking signals that describe external world states while fully accounting for the temporal dimension of the process.

The present work makes the next step showing how the learnt external world dynamics model can be used for prediction of subsequent world states that is important for planning the actions necessary to reach a target world state.

In the next sections we (1) formalize the RL problem solved; (2) describe the neuron model; (3) consider the SNN used for learning world dynamics mode (very briefly); (4) describe the SNN used for inference (which is the main subject of the present article); (5) present the results obtained by our inference-regime network in comparison with the exact external digital inference mechanism. Naturally, this comparison is made on the same the simple but realistic "ping-pong" ATARI RL problem which was used for learning.

## 2 Discretized External World Dynamics

In the present paper, an SNN is considered which includes a model of the external world dynamics in the form of smaller SNN and uses it for prediction of subsequent world states starting from the current one. This smaller SNN was trained inside another SNN structure described in [1] (it will be very briefly considered below). In order to explain how the world dynamics model is expressed by the means of SNN, let us discuss how we formalize it.

We consider the external world state as a Cartesian product of sets of mutually exclusive *elementary states*. We use a discretized representation of the world dynamics, but make a step beyond the classic Markov formalism introducing the temporal coordinate (although, in the discrete form, too). Namely, we formalize the current world state as a set of pairs of integers $\{a \mid <s_{ai}, t_{ai}>\}$, $1 \le s_{ai} \le S_a$, $1 \le t_{ai} \le T_a$, $1 \le a \le S$ where $S$ is the number of elementary components of the whole world state (or elementary state sets - we will also call them dimensions), $S_a$ – the number of elementary states in the set $a$. The second members of the pairs $t_{ai}$ describe how long ago the dimension $a$ of the world state became equal to its present value $s_{ai}$. The positive half of the temporal axis is divided to $T_a$ intervals. The current value $t_{ai}$ indicates that the time elapsed since the moment when the $a$-th dimension became equal to $s_{ai}$, falls within the $t_{a,i}$-th time interval. In this paper, all $T_a$ are equal: $T_a = T$.

Let us explain this world state encoding scheme on the example used in Section V of this article – the simplified model of ping-pong game where the agent controls a racket moving along the left side of the rectangular area in which the ball moves reflecting from three other walls (Fig. 1). The goal – to move the racket so that the ball would hit it keeping it inside the rectangle. This simple world is described by 5 dimensions ($S$ = 5): the 2D ball coordinates, the 2D ball velocity components and the Y coordinate of the racket. Discretized values of these 5 real numbers are $s_{ai}$. Depending on spatial and temporal discretization and the ball speed, for the ball moving to the right far from the right wall, the next elementary state corresponding to the X coordinate after $<x, t>$ may be either $<x + 1, t>$ or $<x, t + 1>$.

The network should learn a predictive model selecting the next most probable world state following the current world state.

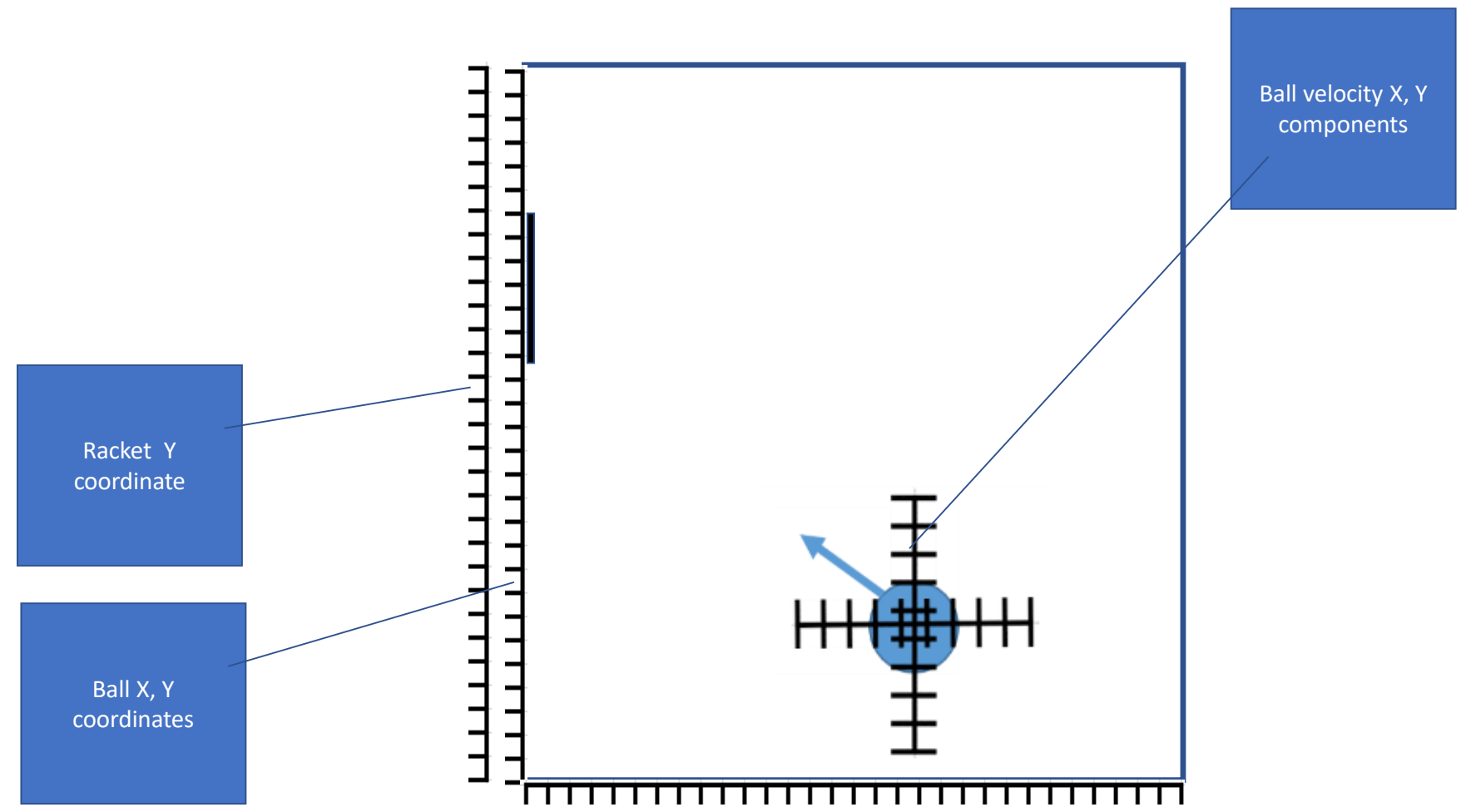

**Fig. 1.** The ping-pong game and the signals fed to the SNN controlling the racket – the coordinates and velocity components of the ball and the Y coordinate of the racket.

# 3 Neuron Model

In this research, we use the LIF (leaky integrate-and-fire) neuron model, which is a simple and widely used neuron model. Additionally, it is implemented—sometimes in a linearized variant—in many modern neurochips, such as Loihi [14], TrueNorth [15], and AltAI [16]. Furthermore, analog implementations of the LIF model also exist [17].

## 3.1 LIF Neuron

The simplest current-based delta synapse model is used for all synapses. It means that every time the synapse receives a spike, it instantly changes the membrane potential by the value of the synaptic weight (it may be positive or negative). Thus, the state of a neuron at the moment t is described by its membrane potential *u(t)*:

$$\frac{du}{dt} = -\frac{u}{\tau_v} + \sum_{i,j} w_i \delta(t - t_{ij}) + I_{noise} \quad (1)$$

and the condition that if $u$ exceeds the threshold value $h$ then the neuron fires and value of $u$ is decremented by $h$. The meaning of the other symbols in (1) is the following: $\tau_v$ – the membrane leakage time constant; $w_i$ - the weight of $i$-th synapse; $t_{ij}$ - the time moment when $i$-th synapse received $j$-th spike. The neuron may have source of its internal stochastic stimulation $I_{noise}$. It is implemented as a random number uniformly distributed in the range [0, $s_{noise}$] which is added to $u$ every simulation step.

Besides that, neuron can have very strong excitatory synapses which force the neuron to fire unconditionally whenever they receive a spike. This firing is called *forced*; after it, the membrane potential is reset to 0.

It should be noted that firing not always causes postsynaptic spike emission (see the next subsection).

The synapses which change the membrane potential may be plastic. Setting the correct values of plastic synapse weights is the aim of the learning process.

### 3.2 Gating Synapses and Neuron Inactivation

A neuron may be in an active or inactive state. If it is inactive, its firing does not cause postsynaptic spike emission. The current neuron regime is determined by the sign of the neuron state component called the *activity time a*. The neuron is active if $a > 0$. Every simulation tact, $a$ changes in accordance with the following formula:

$$a \leftarrow \begin{cases} a + 1 \; if \; a < -1 \\ +\infty \; if \; a = -1 \\ 0 \; if \; a = 0 \\ a - 1 \; if \; a > 0 \end{cases}. \tag{2}$$

Neurons may have special *gating* synapses. A spike coming to a gating synapse with the weight ω changes $a$:

$$a \leftarrow \begin{cases} \min(a, \omega) \; if \; \omega < 0 \\ \max(a, \omega) \; if \; \omega > 0 \end{cases}. \tag{3}$$

Gating connections can be reflective. Reflective blocking connection ($\omega < 0$) is equivalent to refractory period of the length -ω. Reflective activating connection provides a neuron with the ability to repeat the input spike sequence but only while the inter-spike interval in this sequence does not exceed ω.

### 3.3 Emitting Spike Trains

In order to provide neurons with short-term memory mechanism, we introduced the ability of a neuron to emit series of spikes instead of a single spike. It is controlled by a special neuron parameter. If this parameter is greater than 1 then it determines the length of the spike train emitted after neuron firing. This permanent spike emission can be terminated earlier by a spike coming via a gating synapse with negative weight.

### 3.4 Synaptic Plasticity

The SNN considered in this paper is not plastic. Plasticity is utilized in the learning process. When the learning phase is over, in the inference phase, synaptic weights are fixed. Nevertheless, we say some words about our plasticity model. It includes three components:

- **1-factor plasticity.** Every time a plastic synapse receives spike, it weight is slightly decreased.
- **2-factor plasticity** (anti-Hebbian plasticity). When a neuron fires, all its synapses which recently obtained spike are depressed.

- **3-factor plasticity** (dopamine plasticity). Neurons may have special synapses called dopamine synapses. When a spike comes at such a synapse, if the neuron emitted spikes recently, then weights of all its plastic synapses obtaining spikes just before that are changed by the weight of the dopamine synapse.

Besides that, the weight redistribution mechanism works – when some synapse is potentiated, all the rest are depressed slightly – so that the total synaptic weight of one neuron is constant.

# 4 The Learning Network

## 4.1 Representation of Time – the Network Indicating how Long Ago the Current Elementary State was Reached

Now, let us consider how time is represented in our network. Time in its discretized form appears as seconds members of pairs representing the current world state which were introduced in Section 2. In the described learning SNN, the input nodes correspond to elementary states. Thus, it has $\sum_a S_a$ input nodes in total. Constant activity of an input node indicates that the environment is currently in the respective elementary state $s_{ai}$. In order to include time in the internal world model, the network should know how long the world remains in this state – i.e. to determine the respective $t_{ai}$. This problem is solved by a special network module consisting of an array of small SNNs – one SNN per an elementary state (Fig. 2). Each SNN has one input and $T$ output neurons – each output neuron corresponds to one value of $t_{ai}$. The current $t_{ai}$ value is indicated by a spike train emitted by the respective output neuron.

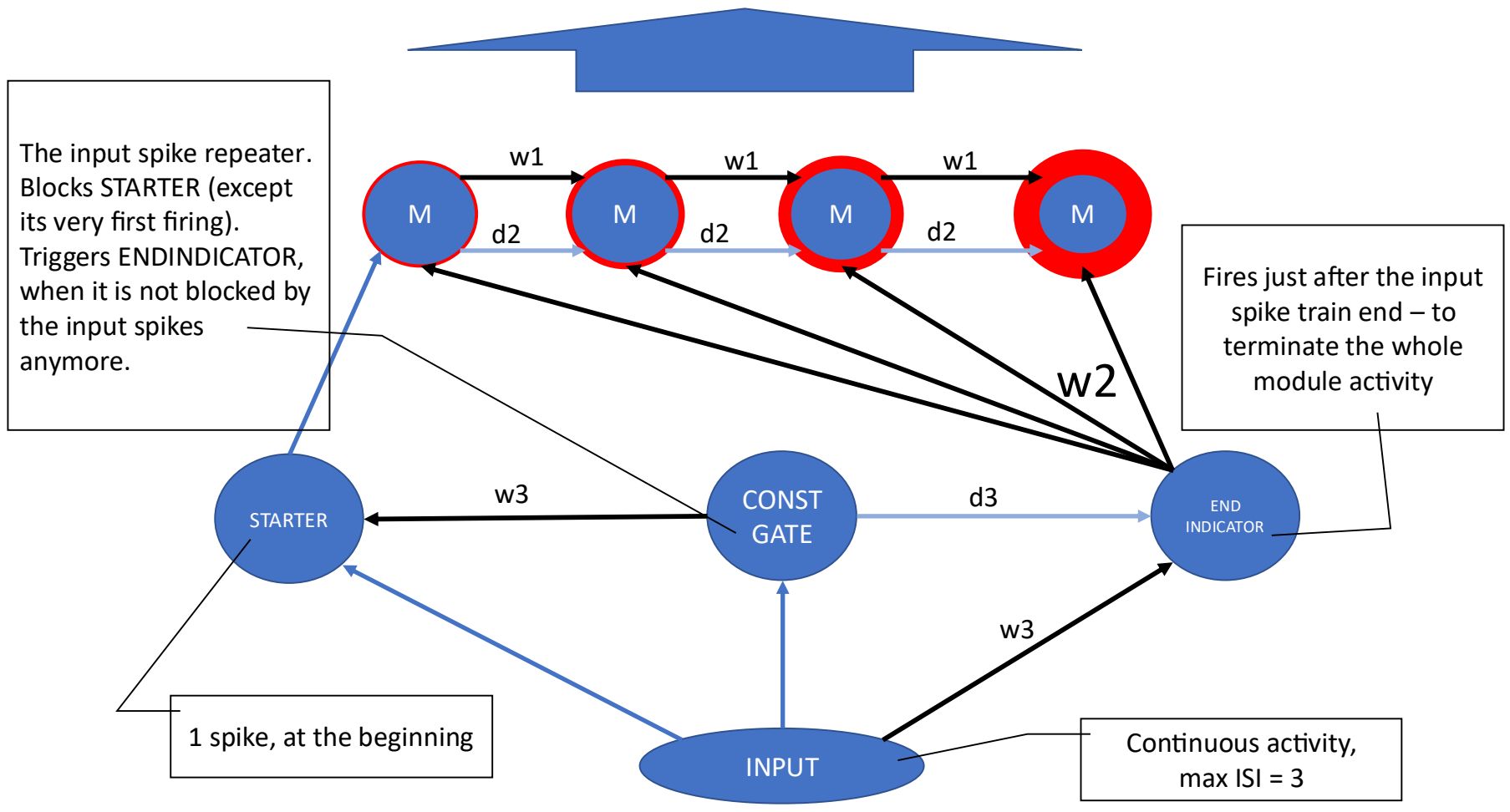

**Fig. 2.** The network module determining how long the input stimulation lasts. Blue arrows are fast forcing connections; black arrows – blocking connections (labelled by their weights); light blue arrows - slow forcing connections (delays are denoted by D). Red bounds denote spike train generation ability. “w” means weight, “d” means delay.

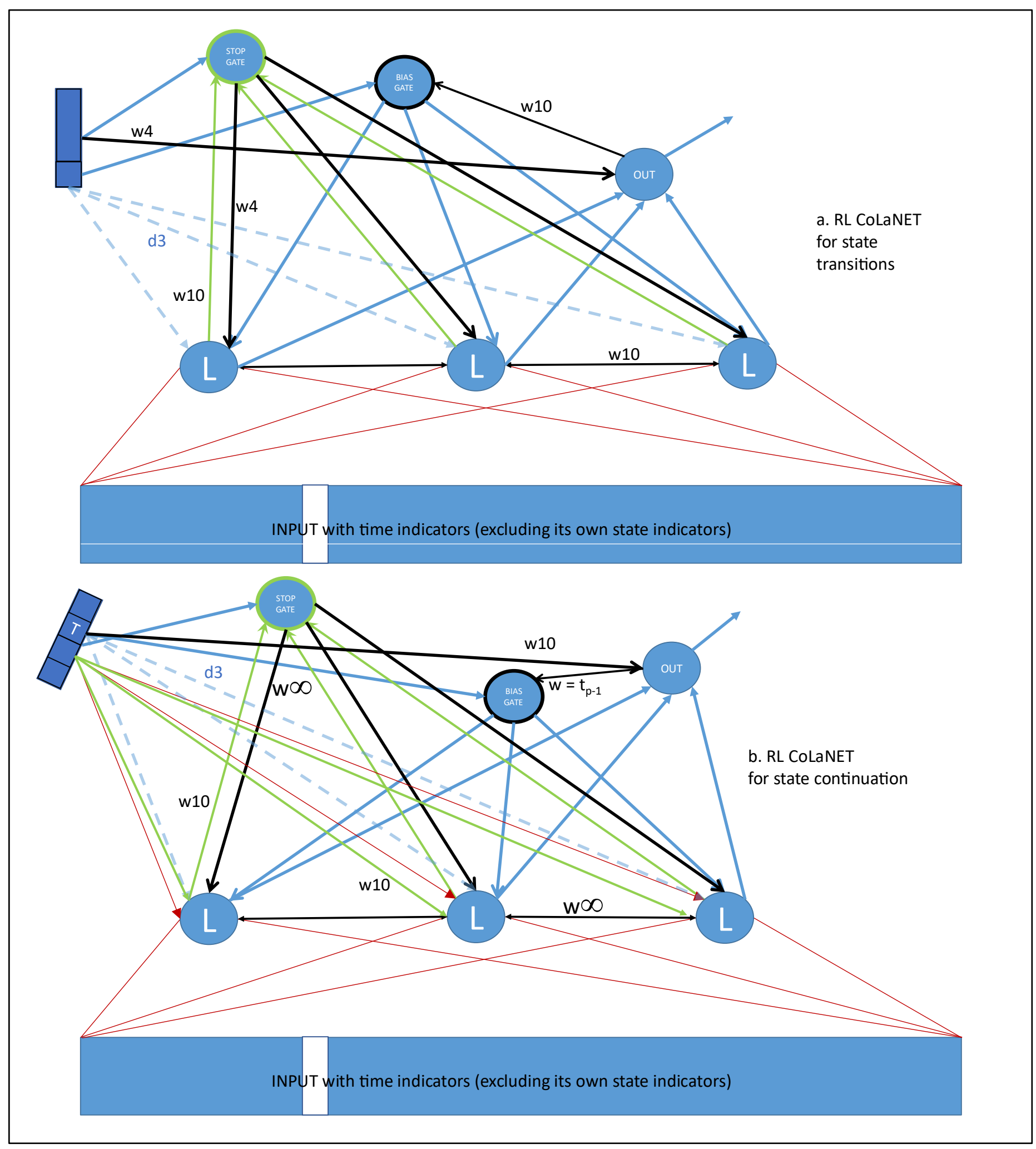

**Fig. 3.** The CoLaNET modifications leveraged for prediction of transition to the new elementary state <$s_{ai}$, *0*> (a) or remaining in the same elementary state – i.e. transition from <$s_{ai}$, *p - 1*> to <$s_{ai}$, *p*> (b). The connection type encoding is the same as for Fig. 2. The dashed arrows correspond to dopamine synapses, the green arrows – to activating connections. The inputs corresponding to $s_{ai}$ are shown separately. On (b), the input corresponding to the target time period is denoted by T.

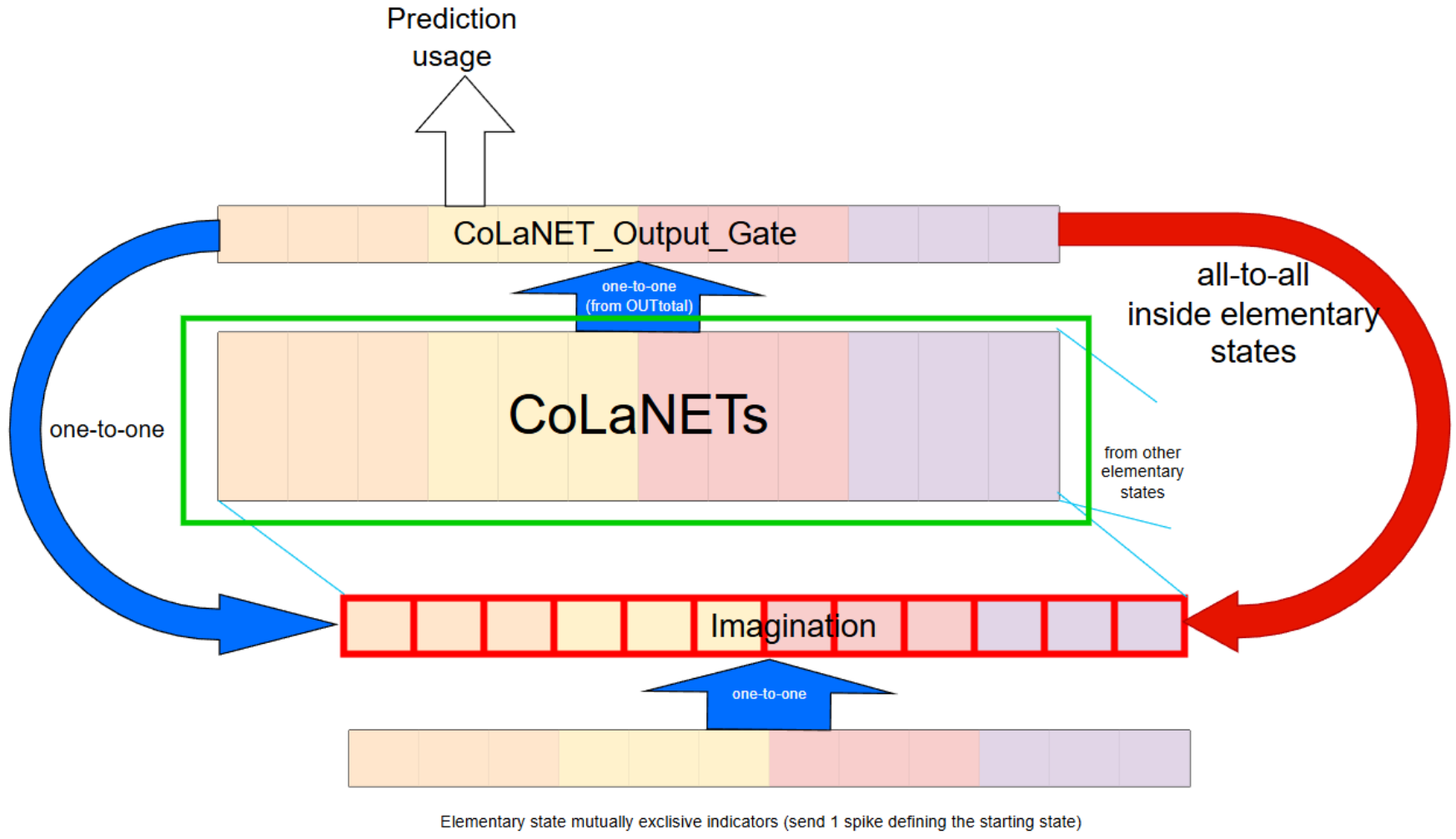

**Fig. 4.** TheSNN structure used for inference. The imported part (the external world dynamics model) is in the green rectangle. The blue arrows are forcing connections; the red arrow is inhibitory connections.

### 4.2 The Learning Subnetwork

The learning subnetwork is an ensemble of CoLaNET classifiers [18 - 20]. One classifier corresponds to one pair $<s_{ai}, t_{ai}>$. Strictly speaking, the classifiers are some modifications of standard CoLaNET, different for zero and non-zero $t_{ai}$. They are schematically depicted on Fig. 3. Every classifier solves the problem of finding the world states from which the world can immediately pass to the given state. The output neurons of the network on Fig. 2 serve as inputs for this learning SNN (therefore, it has $T \sum_a S_a$ inputs). In our study, we exported this part of the whole learning SNN and transferred it into a new SNN designed only for inference. This SNN will be considered in detail in the next Section.

## 5 Testing the SNN Inferring External World Dynamics Rules on the ATARI Ping-Pong RL Problem

The scheme of the SNN used for inference (where the world dynamics model in the form of CoLaNET ensemble was imported) is shown on Fig.4. The model is in the green rectangle. The SNN shown on Fig.4 is used for one time prediction only, but it illustrates the idea which can be used and for permanent predictions, as well. Assume, that it is necessary to predict the sequence of world states after a certain initial state. This initial world state is defined by the input node block at the bottom of Fig. 4. This block has the form and the meaning identical to the inputs on Fig.3. Each of its input nodes emits at most 1 spike defining initial $\{<s_{ai}, t_{ai}>\}$. They send them to the block

"Imagination" which has exactly the same form as the input node block. All neurons in Imagination have memory – once triggered, they continue to emit spikes until they are blocked. The Imagination block serves as input for the CoLaNETs containing the world dynamics model. Some CoLaNET OUT neuron fires predicting the next world state. This spike is repeated by CoLaNET_Output_Gate, used to terminate the prediction process when necessary. The spike denoting a new world state triggers the corresponding neuron in Imagination and, at the same time, it inhibits all other Imagination neurons which correspond to the same world state dimension (to exclude ambiguity). The loop is closed – the network sequentially predicts next world states.

In [1], we evaluated the external world dynamics model obtained for the "ATARI ping-pong" benchmark in the special environment where the logics similar to the described above is hard-coded in a C++ program. Our goal is to reproduce the results reported in [1] but using SNN structures only (not evaluating the model itself).

We preformed the same measurements as in [1] and obtained the following results.

### 5.1 General Accuracy.

The task was to predict at what Y coordinate the ball will hit the left game field bound (where the racket moves). Respectively, the main accuracy measure is standard error of this prediction and the percent of cases when the prediction is not obtained (due to an endless loop in the predicted state sequence). The results are compared in Table 1.

**Table 1.** Accuracy of the Y coordinate prediction for the point where the ball intersects the left game field boundary

| **Inference mechanism** | **Standard error** | | **No-prediction, %** | |
|---|---|---|---|---|
| | ***mean*** | ***standard deviation between series*** | ***mean*** | ***standard deviation between series*** |
| SNN | 8.93 | 0.45 | 5.57 | 1.2 |
| Algorithmic | 8.3 | 0.4 | 3.9 | 1.3 |

We see that the SNN inference implementation is slightly worse, but the difference is not significant (< 2σ).

### 5.2 Dependence of Accuracy of the Left Game Field Boundary Hit Point on the Starting Horizontal Position of the Ball.

Dependences of accuracy of the left game field boundary hit point on the starting horizontal position of the ball for SNN inference and algorithmic inference are shown on Fig. 5. It is seen, that both curves are very similar.

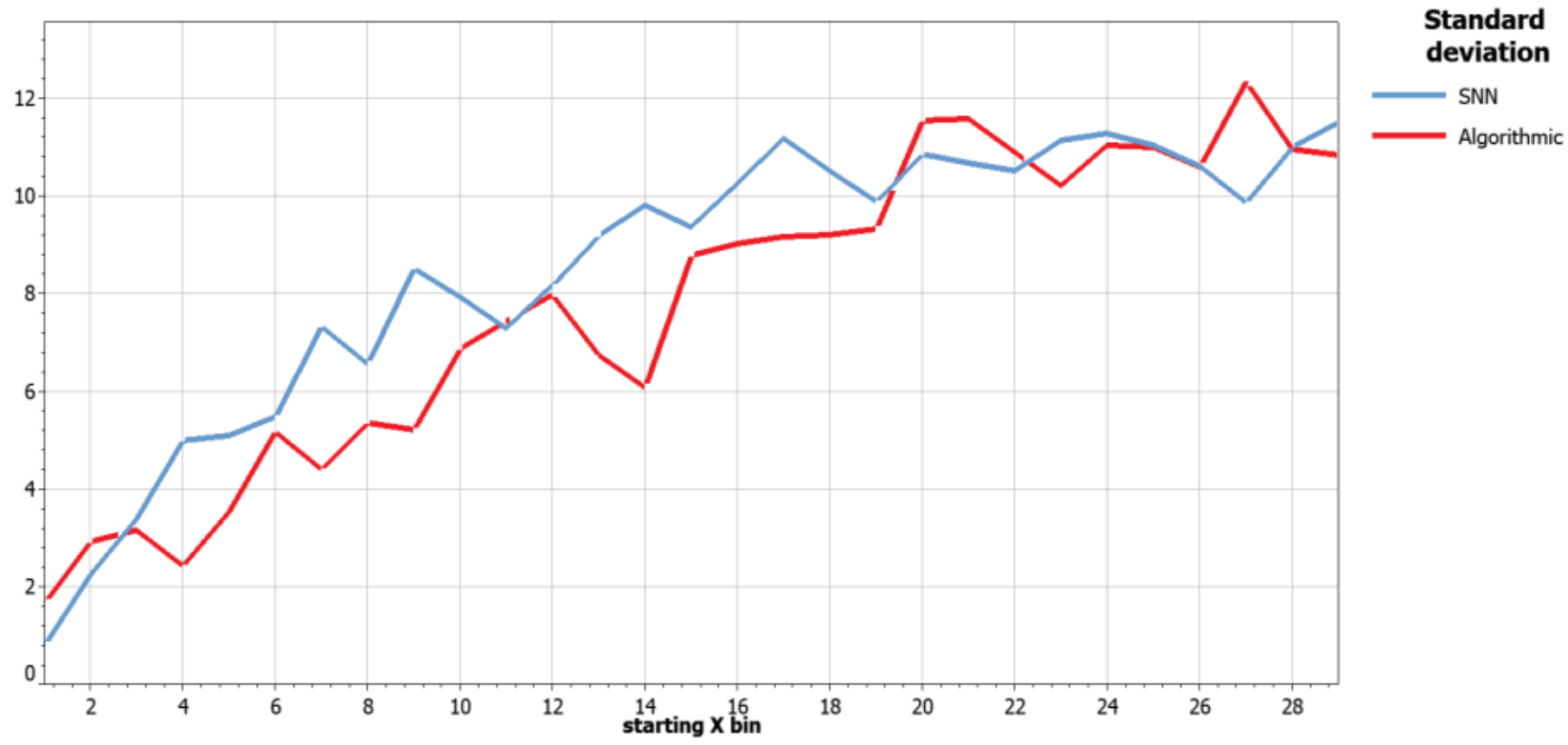

**Fig. 2.** Dependences of accuracy of the left game field boundary hit point on the starting horizontal position of the ball for SNN inference and algorithmic inference.

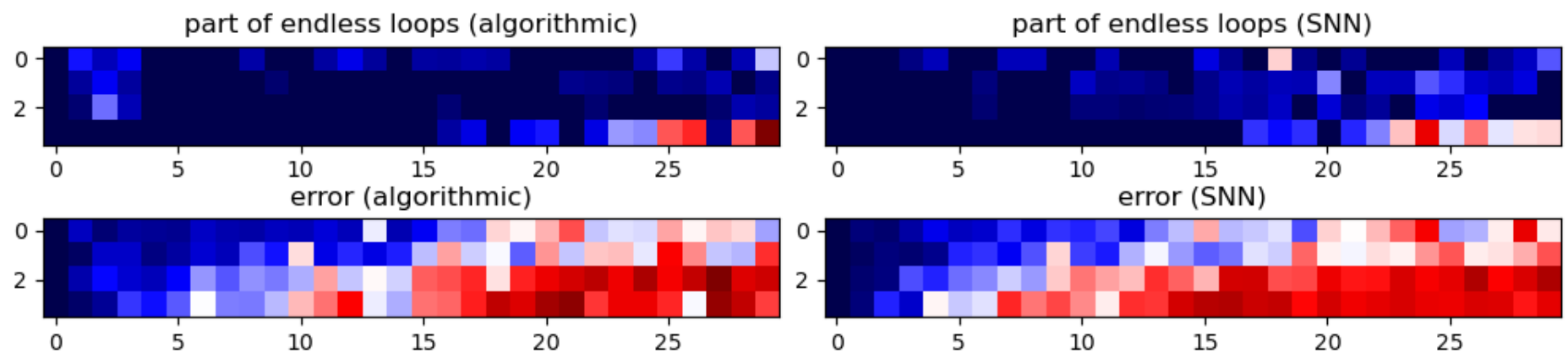

**Fig. 3.** Heatmaps of Dependences of no-prediction case fraction and intersection Y prediction error on starting X bin (the horizontal axes) and starting velocity X component bin (the vertical axes).

### 5.3 Dependence of no-Prediction Case Fraction and Intersection Y Prediction Accuracy on Starting X Bin and Starting Velocity X Component Bin.

The comparison is on Fig.6. Again, the results are very close.

## 6 Conclusion

In this work, we considered the SNN structure incorporating an ensemble of CoLaNET networks which expresses a model of the external world dynamics. This model has been created in another SNN context, and inserted into the proposed SNN for making predictions of subsequent world states starting from the current one.

Before, this was done using an algorithmic (non-network) environment, where all procedures supporting generation of these predictions were implemented as C++ functions. However, efficient implementation of model-based RL on neurochips requires that all its logical components, all related operations would be performed by the respective SNN structures (the “only-spikes, only-SNN” principle), and the present works is a significant step in this direction. The correctness of the SNN implementation of world dynamics model inference is confirmed by the demonstrated proximity of results obtained by the SNN and the algorithmic implementation of the inference mechanism.

The next very important problem to be solved on the way to purely spiking model-based RL is development of an SNN architecture capable of implementing two functions simultaneously – learning and inference, without physical transfer of the learnt model to anywhere. It is the subject of our current research.

## Acknowledgment

The present work is a part of the research project targeted at implementation of model-based reinforcement learning in SNN carried out by Chuvash State University. The author's proprietary SNN simulator package ArNI-X [21] was used to obtain all results reported in this paper.